\documentclass{article}
\usepackage[margin=1.2in]{geometry}
\usepackage{latexsym,verbatim,amsmath,amssymb,amsthm,graphicx,url,color,subcaption,authblk}
\usepackage[round]{natbib}
\usepackage{willcommands}

\DeclareMathOperator* \minn{min\,}
\newcommand{\sbt}{\quad\text{s.t.}~~}
\DeclareMathOperator*\mini{minimize\,}
\DeclareMathOperator*\argmin{argmin}

\newcommand{\rnk}{\text{rank}}

\newtheorem{theorem}{Theorem}

\newtheorem{proposition}[theorem]{Proposition}

\newcommand{\tY}{\widetilde{Y}}
\newcommand{\tU}{\widetilde{U}}
\newcommand{\tV}{\widetilde{V}}
\newcommand{\tX}{\widetilde{X}}

\newcommand{\tGamma}{\widetilde{\Gamma}}

\newcommand{\hY}{\widehat{Y}}

\newcommand{\cL}{\mathcal{L}}

\begin{document}

\title{Flexible Low-Rank Statistical Modeling with Side Information}

\author{William Fithian}
\affil{Department of Statistics, University of California Berkeley}
\author{Rahul Mazumder}
\affil{Sloan School of Management and Operations Research Center, Massachusetts Institute of Technology}

\maketitle

\begin{abstract}
  We propose a general framework for reduced-rank modeling of matrix-valued data.
  By applying a generalized nuclear norm penalty we can directly model
  low-dimensional latent variables associated with rows and columns.
  Our framework flexibly incorporates row and column features,
  smoothing kernels, and other sources of side information by
  penalizing deviations from the row and column models.  Moreover, a
  large class of these models can
  be estimated scalably using convex optimization.  The computational
  bottleneck in each case is one singular value decomposition per iteration
  of a large but easy-to-apply matrix.  Our framework generalizes
  traditional convex matrix completion and multi-task learning methods as well 
  as maximum {\em a posteriori} estimation under a large class of popular hierarchical
  Bayesian models.
\end{abstract}

\section{Introduction}

Matrix completion via low-rank matrix modeling is a central problem in modern multivariate statistics and machine learning. This is due in large part to the advent of collaborative filtering and recommender systems~\citep{agarwal2015statistical}, but matrix completion and matrix factorization applications extend to fields as diverse as image  processing~\citep{angst2011generalized}, X-ray crystallography~\citep{candes2015phase}, seismology~\citep{yang2013seismic}, and political science~\citep{martin2002dynamic,gerrish2011predicting}.

In matrix completion, we partially observe a response matrix $Y\in\R^{n\times m}$,
where each entry $Y_{ij}$ represents a binary, categorical, or real-valued outcome. Typically each row and each column represents an entity of interest such as a user, a test question, an advertisement, or a point in time, and some interaction between the $i$th row entity and the $j$th column entity produces the $Y_{ij}$. In matrix completion problems, only a sparse subset of entries $\Omega \sub \{1,\ldots,n\} \times \{1,\ldots,m\}$ are observed, and the analyst's goal is to predict the missing entries as accurately as possible. To make progress, a common modeling assumption is that each row and column entity can be represented in a latent space of dimension $r \ll n, m$. If $u_i,v_j\in \R^r$ are the latent representations of row $i$ and column $j$ respectively, then any row-column interaction between $i$ and $j$ is assumed to depend only on the inner product $u_i'v_j$. Matrix completion has attracted a great deal of attention in the machine learning community in recent years; see e.g.~\citet{candes:recht,candes2010power,candes2010matrix,keshavan2010matrix,mazumder2010spectral}. A parallel line of work in the multivariate statistics literature dicusses iterative methods to deal with missing values encountered in classical tasks such as principal components analysis~\citep{josse2012handling}, correspondence analysis~\citep{de1988correspondence} multiple correspondence analysis, and multivariate analysis of mixed data sets~\citep{audigier2016principal}. 

Matrix completion became a major focal point of methodological and applied machine learning 
research in part due to the famous Netflix Prize competition of 2006--2009
\citep{bennett2007netflix, bell_koren_07, netflix}, in which researchers competed to improve on Netflix's algorithm for recommending movies to users. In the Netflix data, row $i$ is a particular Netflix user and column $j$ is a movie, and the observed entries are movie ratings from 1--5. $Y_{ij}$ is observed if user $i$ has assigned a rating to movie $j$; otherwise we can interpret it as the rating that user $i$ {\em would assign} to movie $j$ if she were to watch and rate it. Among approximately 480K users and 20K movies, on average each user rates only 200 movies leading to $|\Omega| \approx 10^8$ observed entries out of $nm \approx 10^{10}$ total entries. An objective in such movie-recommender systems is to  recommend movies that their users would enjoy.

As we discuss in Section~\ref{sec:classical}, many of the most familiar algorithms in classical multivariate statistics including principal components analysis, reduced-rank regression, multidimensional scaling, canonical correlation analysis and correspondence analysis can be viewed as estimating lower-dimensional latent factors $u_i$ and $v_j$ to approximate a fully observed matrix under a generalized least-squares loss criterion. 

These classical methods can flexibly incorporate diverse types of side information about row and column entities by imposing constraints or quadratic penalties on the latent factors, with the (truncated) singular value decomposition (SVD)~\citep{GVL83} providing a generic tool for optimizing quadratic objectives with rank constraints. However, if either (a) the matrix is not fully observed or (b) the objective is not quadratic, this framework breaks down. 

Our main aim in this article is to explore an alternative computational and modeling framework, based on convex optimization with a {\em generalized nuclear norm} penalty, that broadens the scope of the SVD framework and adapts it to the more general setting where some entries are missing or the loss function is not quadratic. Section~\ref{sec:lowrank} reviews low-rank approximation and the relationship between the singular value decomposition, the nuclear norm and quadratic regularization in the row and column latent factors, and shows how we can flexibly impose modeling assumptions on the latent variables in matrix completion problems. Section~\ref{sec:computation} reviews algorithmic options and proposes a simple and tractable computational framework based on proximal gradient descent.

Despite their advantages, convex optimization approaches are often dismissed in the collaborative filtering  literature as impractical for large problems since their worst-case scaling is poor and vanilla implementations are not practical; see e.g. \citet{salakhutdinov2008probabilistic,menon2010log}. However, despite the poor worst-case performance, there are rich classes of models in which a generalized version of nuclear norm regularization scales well with problem size, as we will see in Section~\ref{sec:computation}.

Finally, to predict missing data as well as possible, we must understand the {\em missing data mechanism}; that is, we must understand why the data are missing. While the missing-data mechanism has received comparatively little attention in the machine learning literature on matrix completion, the winning team in the Netflix prize reported they saw a breakthrough in their prediction error once they appreciated that users are not selecting movies wholly at random, and the choice of which movies to watch may be quite informative about the user's latent type \citep{bell2007lessons}. By contrast, there is a greater awareness of the missing-data mechanism in the multivariate statistcs literature; see~\citet{audigier2016principal} for a discussion. Section~\ref{sec:missing} discusses how assumptions about the missing-data mechanism can be incorporated into our low-rank modeling framework and how they may complicate the interpretation of our predictions.

\subsection{Preliminaries and Notation}\label{sec:preliminaries}

For a generic matrix $A$, we will generically denote the $i$th row with a lower-case letter $a_i$. We denote $1_n$ as the length-$n$ vector with 1 in every coordinate, and for a matrix $A$ we write the {\em Moore-Penrose psuedo-inverse} as $A^+$. Multiplying $AA^+$ we obtain the projection matrix into the column space of $A$, which we will denote as $\Pi_A$. We assume without loss of generality that $n \geq m$. 
We write the singular values of a matrix $A$ as $\sigma_{1}(A), \ldots, \sigma_{m}(A)$, 
arranged in decreasing order. Let $\|A\|_F^2 = \Tr(A^TA) = \sum_{ij}A_{ij}^2 = \sum_i \|a_i\|_2^2$ denote the {\em Frobenius norm} of matrix $A$.
Let $\|A\|_* = \sum_{k=1}^m \sigma_k(A)$ denote the {\em nuclear norm} of $A$, or the sum of its singular values. By contrast $\|A\|_F^2$ is equal to the sum of its squared singular values, and $\rnk(A)$ is the number of nonzero singular values. In this sense, the rank, nuclear norm, and Frobenius norm are natural matrix counterparts of the $\ell_0$, $\ell_1$, and $\ell_2$ norms for vectors.

\paragraph{Organization of Paper}
Section~\ref{sec:modeling} presents a unified framework of modeling with low-rank and nuclear norm regularization for several problems in classical multivariate statistics and modern collaborative filtering applications -- we also draw parallels with Bayesian modeling schemes.
Section~\ref{sec:missing} presents a connection between the classical missing data literature in statistics and the matrix completion framework.
Section~\ref{sec:computation} presents a broad overview of optimization algorithms that can be used for the class of problems studied herein.
Section~\ref{sec:examples} presents some illustrative examples.

\subsection{Low-Rank Approximation and the Nuclear Norm}\label{sec:lowrank}

Our mathematical point of departure is the prototypical problem of {\em low-rank approximation} to a matrix $Y$ in $\R^{n\times m}$, wherein we attempt to find $r$-dimensional row and column variables $u_i, v_j \in \R^r$ such that $Y_{ij} \approx u_i'v_j$. In matrix notation, $Y\approx UV'$ for $U\in\R^{n \times r}, V\in\R^{m \times r}$, where the quality of approximation is generically measured in terms of some loss function $\cL(\cdot;Y)$:
\begin{equation}\label{eq:rankUV}
  \mini_{U\in \R^{n \times r}, V\in\R^{m\times r}} \cL(UV'; Y).
\end{equation}

Note that in~\eqref{eq:rankUV}, the row and column variables $u_i, v_j$ are intrinsically unidentifiable: for any invertible matrix $A\in \R^{r\times r}$ we could replace $U$ with $\tU = UA'$ and $V$ with $\tV = VA^{-1}$. Then $\tU{\tV}'= UV'$, leading to the same loss. To eliminate this ambiguity, we can introduce the optimization variable $\Theta = UV'$, constraining its rank to be at most $r$, leading to
\begin{equation}\label{eq:rankTheta}
  \mini_{\Theta\in \R^{n \times m}} \cL(\Theta; Y) \sbt \rank(\Theta)\leq r.
\end{equation}

Certain  constraints or penalties that we might impose on $U$ or $V$ translate to constraints or penalties on $\Theta$. For example, in reduced-rank regression we have a feature matrix $X\in\R^{n\times d}$ and require $U=XB$ for some $B\in\R^{d\times r}$; in terms of $\Theta$, it is equivalent to require that $\Pi_{X}^\perp\Theta = 0$. Once we find the best fitting $\Theta$, decomposing it into row variables $U$ and column variables $V$ is essentially a matter of interpretation; if prediction is our aim, it is enough to estimate $\Theta$.

If $\cL(\Theta; Y)$ is the simple least squares loss $\|Y-\Theta\|_F^2$, it is well known that \eqref{eq:rankTheta} is solved by the rank-$r$ truncated SVD~\citep{GVL83} of $Y$. That is, $\Theta = U^rD^r{V^r}'$, where the columns of $U^r\in\R^{n\times r}$ and $V^r \in \R^{m \times r}$ consist, respectively, of the first $r$ left and right singular vectors of $Y$, and $D^r\in \R^{r\times r}$ is diagonal with $D^r_{kk} = \sigma_k(Y)$. If the eigenvalues of $Y$ are all distinct, then the decomposition is unique for every $r$, up to reversing signs of the columns of $U^r$ and $V^r$. As we will see in Section~\ref{sec:classical}, the truncated SVD provides a powerful computational framework for incorporating flexible side information and modeling assumptions about the row and column variables.

Unfortunately, if $\cL$ departs from the Frobenius norm, then~Problem~\eqref{eq:rankTheta}, a non-convex problem owing to the non-convexity of the rank constraint, is generally computationally intractable. In matrix completion, we can only measure errors on the observed entries; thus the natural least-squares loss is $\cL(\Theta; Y) = \sum_{(i,j)\in\Omega}(Y_{ij}-\Theta_{ij})^2$. This seemingly minor variant of the Frobenius norm makes~Problem~\eqref{eq:rankTheta} impossible to solve using a simple SVD. In addition to matrix completion, non-Frobenius loss functions are strongly motivated in several other settings including:
\begin{enumerate}
\item {\bf Sparsely Observed Functional Data:} We are given noisy observations $Y_{ij} = g_i(t_{ij}) + \epsilon_{ij}$ for a smooth function $g_i$ at various locations in a temporal, spatial or other domain, and we wish to reconstruct the functions $g_i$ at unobserved locations. $t_j$ could be values on a real interval or can correspond to spatial locations. This setting is similar to matrix completion, but with notionally infinitely many ``column entities'' representing locations in the domain that may not occur at all in the entire data set; as a result smoothness assumptions are crucial to recover identifiability. We revisit this setting in Section~\ref{sec:functional}
\item {\bf Exponential Families:} If $Y_{ij}$ arises from an exponential family model such as binomial, Poisson, etc., we can model its natural parameter $\Theta_{ij}$ as arising from a low-rank model \citep{roweis1998algorithms,srebro05,natis-05}. Further generalizing this approach, \citet{yee2003reduced} suggest {\em reduced-rank vector generalized linear models} (RR-VGLMs) in which $\Theta = XB$ for some observed covariates $X\in \R^{n\times d}$.
\item {\bf Robust Loss Functions:} If some of the $Y_{ij}$ values may be outliers, we may choose to replace the squared error loss with a more robust entry-wise loss function such as the {\em least absolute deviation loss} $\cL(\Theta; Y) = \sum_{ij} |\Theta_{ij} - Y_{ij}|$ or the {\em Huber loss} \citep{huber2011robust} $\cL(\Theta; Y) = \rho_\gamma(\Theta_{ij} - Y_{ij})$ where
  \[
  \rho_\gamma(x) = \begin{cases}
    \frac12 x^2 & |x| \leq \gamma \\
    \gamma( |x| - \frac12 \gamma) & | x | > \gamma
  \end{cases},
  \]
  a quadratic function for small $x$ but linear in the tails. The least absolute deviation loss (formulated differently), paired with the nuclear norm relaxation (discussed below), forms the basis for the celebrated {\em robust PCA} method of \citet{Candes:rpca}.
\end{enumerate}

Although there are many specialized algorithms for finding approximate solutions or local minima to such models, there is no guarantee that we can solve the problem as posed. A well-designed method may find a suitable local solution or saddle point for many problems, but it can be difficult to predict how these specialized algorithms will perform once we modify the problem to incorporate side information.

\subsubsection{Nuclear Norm Regularization}

By analogy to the Lasso~\citep{Ti96} relaxation of sparse regression, \citet{fazel2002matrix} propose a convex relaxation scheme replacing the rank constraint with a constraint on the nuclear norm, leading to the convex Semidefinite Optimization (SDO) problem:
\begin{equation}\label{eq:nucConstrained}
\minn_{\Theta} \cL(\Theta;Y) \sbt \| \Theta\|_{*} \leq \delta,
\end{equation} 
or in Lagrangian form,
\begin{equation}\label{eq:nucLagrange}
\minn_{\Theta} \cL(\Theta;Y) + \lambda \| \Theta\|_{*}.
\end{equation}

Like the Lasso in linear regression, the nuclear norm plays two roles: first, it promotes a low-rank solution by setting many of the singular values of $\Theta$ to zero;  and second, it regularizes the low-rank solution by shrinking the singular values of $\Theta$ towards zero.

If $\cL(\Theta;Y) = \frac{1}{2}\|Y-\Theta\|_F^2$ then the problem is solved by the {\em nuclear-norm thresholding operator} $S_\lambda(Y)$. $S_\lambda(A)$ is defined as $\diag((A_{11}-\lambda)_+,\ldots,(A_{nn}-\lambda)_+)$ if $A$ is a diagonal $n\times m$ matrix, and otherwise $S_\lambda(A) = US_\lambda(D)V'$ where $UDV'$ is the (full-rank) SVD of $A$. Soft-thresholding the singular values leads to low-rank solutions for large values of $\lambda$. That is, rather than directly constraining the rank, we add a penalty term that favors low-rank solutions.

The nuclear norm may alternatively be viewed as a regularization applied to the latent factors $u_i$ and $v_j$, which can be seen from the following identity appearing in~\citet{fazel2002matrix,srebro2005maximum}:
\begin{equation}\label{eq:ridgeIdent}
  \|\Theta\|_* =  \min_{U,V:\;UV^T=\Theta}\frac{1}{2}\|U\|_F^2 + \frac{1}{2}\|V\|_F^2.
\end{equation}

Applying~\eqref{eq:ridgeIdent}, we can rewrite Problem~\eqref{eq:nucLagrange} in terms of the latent factors as a ridge penalty on the latent factors, with penalty parameter $\lambda$:
\begin{equation}\label{eq:map}
\minn_{U\in\R^{n\times m},V\in\R^{m\times m}} \cL(UV';Y) + \frac{\lambda}{2} \|U\|_F^2 + \frac{\lambda}{2}\|V\|_F^2.
\end{equation}
Although $U$ and $V$ in~Problem~\eqref{eq:map} are formally $n\times m$ and $m\times m$ matrices, their solutions are usually low-rank~\citep{burer-05,hastie2015matrix}, with the rank of the solution adapting to the data $Y$. As a result we can represent them as $n\times r$ and $m\times r$ during the optimization procedure with $r \ll \min(m,n)$, potentially reducing the computational cost dramatically and leading to much more interpretable fitted latent factors.

If $\cL(\cdot; Y)$ is a log-likelihood function or can be interpreted as such~\citep{tipping1999probabilistic}, then~Problem~\eqref{eq:map} is a maximum {\em a posteriori} (MAP) estimation criterion with independent Gaussian priors on the entries of $U$ and $V$ \citep{salakhutdinov2008probabilistic,angst2011generalized,menon2010log}. Note that~Problem~\eqref{eq:map} is {\em not} convex in $(U,V)$, owing to the outer product; however, we can solve for $U$ and $V$ by first solving for $\Theta=UV'$ in~Problem~\eqref{eq:nucLagrange} and then factoring into $U$ and $V$.

Interestingly, solutions to Problem~\eqref{eq:nucConstrained} often approximate solutions to Problem~\eqref{eq:rankTheta} quite well. In their seminal work, \cite{candes:recht} and \citet{candes2010power} study the noiseless matrix completion problem showing that the nuclear norm leads to exact recovery of an underlying low-rank matrix under coherence like assumptions on the underlying matrix even when a few entries of the matrix are observed. \citet{candes2010matrix} and \citet{negahban2012restricted} study theoretical properties of the noisy matrix completion problem using nuclear norm regularization.

Theoretical properties of Problems~\eqref{eq:nucConstrained} and \eqref{eq:nucLagrange} for loss functions beyond squared error have been studied by several authors. For example~\cite{davenport20141} study the problem of \emph{one-bit matrix completion}, where the response is binary and the entrywise loss is logistic, with an additional 
$\ell_{\infty}$-norm on the entries of the matrix, and \cite{lafond2015low} study prediction error bounds for matrix completion for exponential family models with a nuclear norm penalty. \cite{carpentier2016adaptive} discuss confidence sets for the low-rank matrix completion problem and \cite{klopp2015adaptive} consider a multinomial matrix completion problem where the observed entries are quantized with a few levels (in their framework the missingness need not be uniform). They study a regularized negative log-likelihood problem, where the latent variables are regularized by a nuclear norm penalty and an additional constraint on the maximal absolute entries of the matrix. \citet{udell2016generalizednew} extend a previous version of this manuscript and discuss computational aspects of low-rank modeling arising in machine learning problems.

\subsubsection{The Generalized Nuclear Norm}

We can generalize the penalty in Problem~\eqref{eq:nucLagrange} to obtain
\begin{equation}\label{eq:genNuc}
   \minn_\Theta \cL(\Theta;Y) + \lambda\|P\Theta Q\|_*
\end{equation}
for positive semidefinite matrices $P$ and $Q$. We can interpret $P$ and $Q$ in~Problem~\eqref{eq:genNuc} as modulating the degree of $\ell_2$ penalization for $U$ and $V$ respectively, by penalizing some directions more than others. For example, if $X\in\R^{n\times d}$ is a matrix of features for the rows then we might use $P= \mathbb{I}-\Pi_X$ (where, $\mathbb{I}$ is the identity matrix) so that the component of $\Theta$ explained by $X$ is unpenalized (see Section~\ref{sec:modelNuc}).

Several other authors have proposed interesting specific applications of the
generalized nuclear norm--\cite{salakhutdinov2010collaborative} advocate a special case of
Problem~\eqref{eq:genNuc} with diagonal $P$ and $Q$,  and \citet{angst2011generalized} apply the generalized nuclear norm to the structure-from-motion problem in computer vision.
\citet{abernethy2009new} frame collaborative filtering in very general terms of estimating compact linear operators in Hilbert space.  Their proposals for regularizing $\Theta$ have the most overlap with ours but with less focus on scalable computation.

Provided that $\cL$ is convex in $\Theta$, Problems \eqref{eq:nucConstrained}, \eqref{eq:nucLagrange}, and \eqref{eq:genNuc} can be solved in polynomial time using standard convex optimization techniques. Convex optimization is appealing because it allows abstraction of our statistical model from our estimation algorithm. Even so, the computational cost of off-the-shelf interior point solvers become prohibitively large as soon as the problem sizes become larger than a few hundred. Towards this end, first order\footnote{First-order optimization algorithms are iterative methods with significantly low per iteration cost when compared to Interior Point algorithms. Even if first order methods take many more iterations than an Interior Point algorithm 
to converge to a solution with comparable accuracy, their low-memory requirement and cheap per iteration cost makes them applicable to modern large scale problem instances. In addition, low to moderate accuracy solutions lead to excellent estimates with good statistical properties especially in large noisy datasets~\citep{boto-08}.} methods~\cite{nesterov2004introductory} are used to obtain low-to-moderate accuracy solutions. Indeed, as we discuss in Section~\ref{sec:computation}, developing fast, scalable and rigorous algorithms continues to be an active area of research, across the fields of statistics, machine learning and optimization.

If $P$ and $Q$ are invertible then a simple change of variables argument shows that we can rewrite~Problem~\eqref{eq:genNuc} as $\smash{\cL(UV') + \frac{\lambda}{2}\|PU\|_F^2 + \frac{\lambda}{2}\|QV\|_F^2}$. In fact, the same holds for generic semidefinite $P$ and $Q$, as we state below and prove in Appendix~\ref{sec:proof-propProj}.
\begin{proposition}\label{propProj}
  Let $P\in\R^{n\times n}$ and $Q\in\R^{m\times m}$ be positive
  semidefinite.  Then for any function
  $\cL:\R^{n\times m}\to \R$, we have
  \begin{align}
  \inf_{U \in \R^{n \times r} ,V \in \R^{m \times r} }&\cL(UV') + \frac{1}{2}\|PU\|_F^2 + \frac{1}{2}\|QV\|_F^2  \label{eq:crit1}\\
    =\,\quad\inf_{\Theta} \quad&\cL(\Theta)
    + \|P \Theta Q\|_* \label{eq:crit0}  \\ 
    =\inf_{\Theta_1,\Theta_2,\Theta_3} &\cL(P^+\Theta_1Q^+
    + \Pi_P^\perp\Theta_2 + \Theta_3\Pi_Q^\perp) + \|\Theta_1\|_*,\label{eq:crit2}
\end{align}
  where, $r=\min \{m ,n \}$ and $P^+$ and $Q^+$ are the Moore-Penrose pseudoinverses of $P$
  and $Q$, and $\Pi_P^\perp$ and $\Pi_Q^\perp$ are projections onto
  their respective null spaces.
\end{proposition}

Proposition~\ref{propProj} is useful because it allows us to move easily back and forth between modeling latent factors via the more interpretable formulations (\ref{eq:crit1}--\ref{eq:crit0}) and the more computationally favorable formulation Problem~\eqref{eq:crit2}. As we will discuss further in Section~\ref{sec:computation}, formulation \eqref{eq:crit2} is often computationally attractive for two reasons. First, we may be able to represent $\Theta_2$ and $\Theta_3$ as matrices of much smaller dimension, while $\Theta_1$ is low-rank and can be represented efficiently as an outer product of smaller matrices. If $P = \Pi_{X}^\perp$, for example, then $\Pi_P^\perp = \Pi_X$, and we can replace the unpenalized term $\Pi_P^\perp\Theta_2$ with $XB$, where $B\in\R^{d \times m}$. Second, if we use proximal gradient descent then the proximal steps for $\Theta_1$ will be with
respect to the standard nuclear norm, so they can be solved in closed
form using a soft-thresholded SVD (by contrast, proximal gradient steps
with respect to a generalized nuclear norm generically cannot be
solved in closed form).

\subsection{Other Approaches}\label{sec:review}

\paragraph{The Max-Norm regularization:}
The max-norm is another convex proxy for the rank of a matrix that are often used in the context of matrix completion and related problems~\citep{srebro2004maximum,srebro2005rank}. 
The max-norm of a matrix $A$ is defined as $\|A\|_{\infty,2} = \max_{i=1,\ldots,n}\|a_i\|_2$. Convex and closely related to the nuclear norm, it can be defined via matrix factorizations as:
\begin{equation}\label{max-norm1}
 \|\Theta\|_{\max}  =  \min_{ U, V: \Theta = UV'} \left(\| U \|_{2, \infty} \| V\|_{2, \infty}  \right).
 \end{equation}
\cite{lee2010practical} demonstrate that the empirical performance of a max-norm regularized version of matrix completion may lead to better predictive performance on some collaborative filtering datasets.  Theoretical properties of the max-norm have been studied by \cite{srebro2004maximum,srebro2005rank,foygel2011concentration,cai2013max}, but max-norm regularization is computationally more challenging than nuclear-norm regularization.

\paragraph{Bayesian Methods:}
Another way to incorporate domain knowledge or side information is to use complex hierarchical Bayes models which can be fit using various specialized approaches. For example, \cite{salakhutdinov2008bayesian} study a generative model with additional priors on the hyper-parameters and develop a Gibbs sampling scheme for the problem, leading to a computationally intensive method requiring 200 hours to train a model with $r=60$ on the Netflix dataset. \citet{agarwal2009regression} also propose a more general Bayesian modeling framework which we revisit later in Section~\ref{sec:modelNuc}. \cite{AZM-11} study an example where these covariance matrices are unknown and they are estimated via inverse covariance matrix estimation. \cite{todeschini2013probabilistic}  place a prior on the singular values of the matrix and propose an EM-stylized algorithm for the task. \cite{cottet20161} study the one-bit matrix completion from a Bayesian perspective using variational techniques.

Because the resulting model specifications are highly non-convex, doing tractable inference or making formal statements about the quality of the estimates obtained are rather challenging. However, as we show in Section~\ref{sec:modelNuc}, our generalized nuclear norm regularization framework can be used to perform MAP inference in these models. Even if our goal is to sample from the posterior rather than find the MAP estimate, it is often practically useful to use optimization  techniques to understand properties of the posterior distribution or help with the sampling. In this vein, \citet{agarwal2009regression} empirically  compare several methods for fitting the same model, and settle on an estimation method of their own devising, called Monte Carlo-EM, to find a local maximum of the marginal likelihood.

\section{Modeling in Latent Space}\label{sec:modeling}

\subsection{Low-Rank Modeling with the SVD}\label{sec:classical}

Some of the most familiar methods in classical statistics amount to low-rank least-squares approximation of an appropriate matrix, along with some preprocessing of the matrix and postprocessing of the singular vectors. We review some examples here, for more details and a classical perspective see \citep{mardia1980multivariate}.

\paragraph{Principal Components Analysis:} 

Principal components analysis (PCA) computes the directions of greatest variation among rows of a data matrix. We begin by subtracting the mean from each column, obtaining $\tY = Y - n^{-1}1_n1_n'Y = \Pi_{1_n}^\perp Y$. The first $r$ principal components and principal component loadings are, respectively, the columns of $U^rD^r$ and $V^r$ where $U^rD^r{V^r}'$ is the rank-$r$ truncated SVD of $\tY$. If $\bar Y = n^{-1}Y'1_n$, the vector of column means, then we can reconstruct a least-squares approximation to $Y$ as $\hY = 1_n\bar Y' + U^rD^r{V^r}'$.

Consider modeling $Y_{ij}\simind N(\Theta_{ij}, \sigma^2)$ where $\Theta_{ij} = \beta_j + u_i'v_j$, or in matrix form $\Theta = 1_{n}\beta' + UV'$ ($1_{n} \in \Re^{n}$ is a vector of all ones) where $\beta\in \R^m, U\in\R^{n\times r}$, and $V\in\R^{m \times r}$. Because $UV'$ can be any matrix with rank less than $r$, we can equivalently write $\Theta = 1_n\beta' + \Gamma$ where $\Gamma\in\R^{n\times m}$ with $\rank(\Gamma)\leq r$.
In this model, the maximum likelihood estimator for $\Theta$ solves
\begin{equation}\label{eq:PCAmod}
  \minn_{\Theta} \|Y - \Theta\|_F^2 \sbt \Theta = 1_n\mu' + \Gamma, \;\;\rank(\Gamma) \leq r.
\end{equation}

Note that for any solution $(\beta,\Gamma)$ to Problem~\eqref{eq:PCAmod} with $1_n'\Gamma \neq 0$ (i.e., some column of $\Gamma$ has nonzero mean), the alternate solution $(\beta + n^{-1}\Gamma'1_n, \Pi_{1_n}^\perp \Gamma)$ leads to exactly the same $\Theta$ value, and hence the same likelihood. Because $\rank(\Pi \Gamma) \leq \rank(\Gamma)$ for any projection matrix $\Pi$, we have no reason to entertain solutions with $1_n'\Gamma \neq 0$. Thus, we can add the constraint $\Pi_{1_n}\Gamma=0$ without changing the estimation problem. As a result we have  $\Pi_{1_n}^\perp \Theta = \Gamma$ and $\Pi_{1_n}\Theta = 1_n\beta'$. 

Eliminating $\beta$ and $\Gamma$ from the problem, we can rewrite it in condensed form as
\begin{equation}\label{eq:PCAopt}
  \minn_{\Theta} \|Y - \Theta\|_F^2 \sbt \rank(\Pi_{1_n}^\perp \Theta) \leq r.
\end{equation}
In other words, the rank constraint only applies to the portion of the column space of $\Theta$ that is orthogonal to $1_n$. In that sense we can say $1_n$ is an unregularized column direction.

Having derived Problem~\eqref{eq:PCAopt} we can easily solve for the maximum likelihood estimator by noting that
\begin{align*}
\|Y-\Theta\|_F^2 
  &= \|\Pi_{1_n}Y - \Pi_{1_n}\Theta\|_F^2 + \|\Pi_{1_n}^\perp Y - \Pi_{1_n}^\perp\Theta\|_F^2\\
  &= \|1_n\bar Y' - \Pi_{1_n}\Theta\|_F^2 + \|\tY - \Pi_{1_n}^\perp\Theta\|_F^2.
\end{align*}
We can set the first term to zero by taking $\Pi_{1_n}\Theta=1_n\bar Y'$ (leading to $\beta=\bar Y$) and minimizing $\|\tY - \Pi_{1_n}^\perp\Theta\|_F^2$ via the SVD (leading to $\Gamma = UDV'$).

\paragraph{Reduced Rank Regression:}  

As a second example, suppose we have a response matrix $Y\in \R^{n\times m}$ and feature matrix $X\in \R^{n\times d}$, and consider regressing each column of $Y$ on the predictors $X$, but sharing information across the $m$ responses via a rank constraint. That is, suppose we again model $Y_{ij}\simind N(\Theta_{ij}, \sigma^2)$, but now modeling $\Theta_{ij} = \alpha_j + x_i'\beta_j$, for $j=1,\ldots,m$, and with a constraint on the rank of $B = [\beta_1 \cdots \beta_m]$, leading to the popular reduced-rank regression model~\citep{anderson1951estimating,velu2013multivariate}. The maximum likelihood problem can then be written as
\begin{equation}\label{eq:RRRmod}
  \minn_{\Theta} \|Y - \Theta\|_F^2 \sbt \Theta = 1_n\alpha' + XB, \;\; \rank(B) \leq r.
\end{equation}

By a similar logic as before, we may assume without loss of generality that the columns of $X$ have mean zero: if $\tX = \Pi_{1_n}^\perp X$ and $(\alpha, B)$ solves Problem~\eqref{eq:RRRmod} with data $(\tX,Y)$ then $(\alpha + n^{-1}(XB)'1_n, B)$ solves Problem~\eqref{eq:RRRmod} with data $(X,Y)$. Furthermore, noting that
\[
\{XB:\; B\in \R^{d \times m}, \rank(B) \leq r\} = \{A\in \R^{n \times m}:\; \Pi_X^\perp A = 0, \rank(A)\leq r\},
\]
we can eliminate $\alpha$ and $B$ and obtain
\begin{equation}\label{eq:RRRopt}
  \minn_{\Theta} \|Y - \Theta\|_F^2 \sbt \rank(\Pi_{1_n}^\perp\Theta) \leq r, \;\; \Pi_{[1_n, X]}^\perp\Theta = 0.
\end{equation}
In Problem~\eqref{eq:RRRopt} we see a similar prioritization of column directions as in Problem~\eqref{eq:PCAopt}, only now with three levels of prioritization: $1_n$ is unregularized, the column space of $X$ is regularized via a rank constraint, and all other directions are completely killed.

As before, we can solve Problem~\eqref{eq:RRRopt} by decomposing the Frobenius norm into the three column spaces of interest:
\begin{align*}
\|Y - \Theta\|_F^2 
  &= \|\Pi_{1_n}Y - \Pi_{1_n}\Theta\|_F^2 + \|\Pi_{X}Y - \Pi_{X}\Theta\|_F^2 + \|\Pi_{[1_n,X]}^\perp Y\|_F^2.
\end{align*}
We can eliminate the first term by taking $\alpha=\bar Y$. To minimize the second term we set $\Pi_X \Theta = XB = UDV'$, the rank-$r$ truncated SVD of $\Pi_X Y$, and solving for $B$ we obtain $B = X^+UDV'$. The third term depends only on $Y$ and does not influence the solution.

\paragraph{Non-Identity Covariance, Row Effects and Further Generalizations:}

The basic formulations of PCA and reduced-rank regression above are natural if the $m$ columns of $Y$ are measured in the same units and errors are of a comparable scale. In other cases, it would be more natural to measure the approximation error relative to a different metric based on the modified log-likelihood. For example, suppose that $Y_{ij} \simind N(\Theta_{ij}, \Sigma)$ for some known or estimated error covariance matrix $\Sigma \succ 0$. 

In that case, the maximum-likelihood problem for PCA becomes
\begin{equation}\label{eq:PCAnonid}
  \minn_\Theta \|(Y-\Theta)\Sigma^{-1/2}\|_F^2 \sbt \rank(\Pi_{1_n}^\perp \Theta) \leq r.
\end{equation}
Making the same decomposition as before, we will set $\beta = \bar Y$ to eliminate the residual in the direction of $1_n$, but to minimize the second term the solution for $\Gamma$ will solve
\begin{equation}\label{eq:PCAnonid:Gamma}
\minn_\Gamma \|\tY\Sigma^{-1/2} - \Gamma\Sigma^{-1/2}\|_F^2 \sbt \rank(\Gamma) \leq r.
\end{equation}

To solve Problem~\eqref{eq:PCAnonid:Gamma} we can simply change variables to $\tGamma = \Gamma\Sigma^{-1/2}$, noting that $\rank(\Gamma)=\rank(\tGamma)$ for any $\Gamma\in\R^{n\times m}$. Then we see the minimizer is $\tGamma=UDV'$, the rank-$r$ truncated SVD of  $\tY\Sigma^{-1/2}$, and $\Gamma=UDV'\Sigma^{1/2}$. Using a further change of variables, we could also handle the more general case where the loss function is replaced by $\|\Phi^{-1/2}(Y-\Theta)\Sigma^{-1/2}\|_F^2$.

Generalizing in another direction, we might wish our model to incorporate row-wise fixed effects in addition to column-wise fixed effects, but with low-rank interactions between rows and columns. In that case, we might model $\Theta_{ij} = \alpha_i + \beta_j + \Gamma_{ij}$, leading to the likelihood criterion 
\begin{equation}\label{eq:PCArowmod}
  \minn_\Theta \|Y-\Theta\|_F^2 \sbt \Theta = \alpha 1_m' + 1_n\beta' + \Gamma, \;\; \rank(\Gamma)\leq r.
\end{equation}
By a similar argument as in the previous section, we can assume without loss of generality that both $1_n'\Gamma=0$ and $\Gamma 1_m=0$, leading to the condensed criterion
\begin{equation}\label{eq:PCArowopt}
  \minn_\Theta \|Y-\Theta\|_F^2 \sbt \rank(\Pi_{1_n}^\perp\Theta\Pi_{1_m}^\perp)\leq r.
\end{equation}
In this case there is an unregularized column direction {\em and} an unregularized row direction; the rank constraint only applies to the portion of the model orthogonal to both. The answer can still be computed in closed form via a truncated SVD of $\Pi_{1_n}^\perp\Theta\Pi_{1_m}^\perp$.

By combining and extending the ideas above, many further generalizations are possible. As long as we use a generalized least-squares loss function of the form $\|\Phi^{-1/2}(Y-\Theta)\Sigma^{-1/2}\|_F^2$, we can choose from a great variety of models for $\Theta$ that are all computable in closed form using a common computational framework based on the SVD. Unfortunately, however, the SVD framework breaks down when we attempt to generalize beyond a fully observed least-squares loss, as we see next.

\subsection{Low-Rank Modeling with Nuclear-Norm Regularization}\label{sec:modelNuc}

When we move from the rank-constrained problem to the nuclear-norm-regularized problem we can use essentially all of the same manipulations as in the previous section to reduce any constraints on $U$ and $V$ to constraints on $\Theta$. In addition to constraining the latent factors, however, we have an additional option to impose Bayesian priors on the factors and fit the resulting models by MAP estimation. We again discuss several examples below.

\paragraph{Matrix Completion:}
In matrix completion problems like the Netflix challenge, a simple and appealing model is to assign a marginal effect to each row and column entity (e.g., a movie's overall quality and a user's overall affect) as well as a low-rank interaction, leading to the model $\Theta_{ij} = \alpha_i + \beta_j + u_i'v_j$. In matrix form, we can write the constraint on $\Theta$ as $\Theta = \alpha 1_m' + 1_n\beta' + UV'$, or equivalently that 
 \begin{equation}\label{eq:Netmod}
   \minn_{\Theta,\alpha,\beta,\Gamma} \cL(\Theta; Y) + \lambda\|\Gamma\|_* \sbt \Theta = \alpha 1_m' + 1_n\beta' + \Gamma.
\end{equation}

We can eliminate $\alpha$ and $\beta$ from the problem using a similar argument as in the last section to rewrite Problem~\eqref{eq:PCArowmod} as \eqref{eq:PCArowopt}: for any solution $(\alpha,\beta,\Gamma)$ with $1_n'\Gamma \neq 0$, the alternate solution $(\alpha, \beta + n^{-1}\Gamma'1_n, \Pi_{1_n}^\perp \Gamma)$ leads to the same loss but a smaller nuclear norm for $\Gamma$, since $\Gamma'\Pi_{1_n}^\perp\Gamma' \preceq \Gamma'\Gamma$ in semidefinite ordering. Making a similar argument for $\alpha$, we can rewrite~Problem~\eqref{eq:Netmod} as
\begin{equation}\label{eq:Netopt}
  \minn_{\Theta} \cL(\Theta; Y) + \lambda\|\Pi_{1_n}^\perp \Theta \Pi_{1_m}^\perp\|_*,
\end{equation}
leading to a well-defined convex optimization problem. The positive-semidefinite matrices $P=\Pi_{1_n}^\perp$ and $Q=\Pi_{1_m}^\perp$ in the penalty encode our decision to include $\alpha_i$ and $\beta_j$ as free parameters; if $\Theta = UV'$ then $U$ and $V$ are only penalized insofar as they deviate from constants.

Note that nothing about the reduction above required any assumption on the form of $\cL$. Hence, we can use the same reduction with an entrywise exponential family likelihood, or Huber or absolute deviation loss; as long as $\cL$ is convex in $\Theta$, we arrive eventually at a convex optimization problem.

\paragraph{Features and Reduced-Rank Vector GLMs:}

Extending the previous model, we might choose to model the row effects as a linear function of the row-feature matrix $X\in\R^{n\times d}$. One option already discussed is to penalize only $\Pi_X^\perp\Theta$, imposing the model $\Theta_{ij} = \alpha_j + x_i'\beta_j + \Gamma_{ij}$ where only the saturated interaction matrix $\Gamma$ is penalized. If $B = [\beta_1 \cdots \beta_m]\in\R^{d\times m}$, this model leads to the criterion
\begin{equation}\label{eq:nucPenRegression}
   \minn_{\Theta,\alpha,B,\Gamma} \cL(\Theta; Y) + \lambda\|\Gamma\|_* \sbt \Theta = 1_n\alpha' + XB + \Gamma,
\end{equation}
which in condensed form is
\begin{equation}
  \minn_\Theta \cL(\Theta; Y) + \lambda\|\Pi_{[1,X]}^\perp \Theta\|_*
\end{equation}
As above, $\Gamma$ is then penalized insofar as it is not explained by the features $X$.

A second option is to constrain $\Gamma = XB$ while using nuclear-norm regularization to enforce that $B$ is (approximately) low-rank, leading to a reduced-rank vector GLM. If we still allow for an unpenalized intercept $\alpha_j$, we could write the problem as
\begin{equation}\label{eq:rrvglm}
  \minn_\Theta \cL(\Pi_{[1_n,X]} \Theta; Y) + \lambda\|\Pi_{1_n}^\perp\Theta\|_*,
\end{equation}
which is always minimized by some $\Theta$ for which $\Pi_{[1,X]}\Theta=\Theta$ (otherwise $\Pi_X\Theta$ would give a smaller nuclear norm without changing the loss). 

Note that if we allow the matrices $P$ and $Q$ to have some infinite eigenvalues, then (abusing notation) we could alternatively write
\begin{equation}\label{eq:rrvglm2}
  \minn_\Theta \cL(\Theta; Y) + \lambda\|(\Pi_{1_n}^\perp + \infty\Pi_{[1,X]}^\perp)\Theta\|_*,
\end{equation}
where the infinite eigenvalues mean only that $\Theta$ is completely disallowed from varying in that direction (more precisely we can imagine~Problem~\eqref{eq:rrvglm2} as a limit of problems with $C\Pi_{[1,X]}^\perp$ replacing $\infty\Pi_{[1,X]}^\perp$, $C\to\infty$). Thus, intercepts are unpenalized, the other directions in the span of $X$ are penalized equally, and directions outside the span of $[1,X]$ are completely killed.

The two solutions discussed above are quite different but both enforce multitiered regularization among different left-directions of $\Theta$. One can imagine an infinite variety of penalization schemes obtained by prioritizing left- and right-directions in the same way.

\paragraph{Priors on Latent Factors and MAP Estimation:}

By interpreting the nuclear norm penalty as a ridge penalty on latent factors, we reformulate MAP estimation in a variety of interesting Bayesian models as convex optimization problems. If $\Sigma\in\R^{n\times n}$ and $\Phi\in\R^{m\times m}$ are positive-definite covariance matrices reflecting correlations between rows of the latent-factor matrices $U$ and $V$, we can impose the Bayesian model:
\begin{equation}\label{eq:simpleBayes}
  \begin{aligned}
  &U_k \simiid N_n(0,\Sigma), \quad V_k \simiid N_m(0,\Phi), \quad &k=1,\ldots,r\\
  &Y_{ij} \mid u_i, v_j \simind N(u_i'v_j, \tau^2),\quad &(i,j)\in\Omega,
\end{aligned}
\end{equation}
where $U_k$ is the $k$th column of $U$ and $u_i$ is the $i$th row. Up to a constant shift, the negative log-posterior is
\begin{align}
  \sum_{(i,j)\in\Omega} \frac{1}{2\tau^2} (Y_{ij}-u_i'v_j)^2 
  &+ \sum_k\frac{1}{2}U_k'\Sigma^{-1}U_k + \sum_k \frac{1}{2}V_k'\Phi^{-1}V_k\\
  \label{eq:simpleBayesFrob}
  &= \cL(UV'; Y) + \frac{1}{2}\|\Sigma^{-1/2}U\|_F^2 + \frac{1}{2}\|\Phi^{-1/2}V\|_F^2.
\end{align}
where $\smash{\cL(UV'; Y) = \sum_{(i,j)\in\Omega} \frac{1}{2\tau^2} (Y_{ij}-u_i'v_j)^2}$. Using Proposition~\ref{propProj} we can rewrite~\eqref{eq:simpleBayesFrob} as
\begin{equation}
  \cL(\Theta; Y) + \|\Sigma^{-1/2}\Theta\Phi^{-1/2}\|_*. 
\end{equation}
As always, this reduction is equally correct if we replace the Gaussian log-likelihood for $Y$ with any other log-likelihood for $Y$ given $UV'$. As long as the log-likelihood is convex in $UV'$, we obtain a convex problem in terms of $\Theta$ (as long as the rank of $U,V$ are sufficiently large). 

\paragraph{MAP Estimation for Hierarchical Priors:}
To incorporate more domain knowledge or side information, various authors have proposed more complex hierarchical Bayes models which they estimate using various specialized approaches.  A general modeling framework to tackle complex problems arising in recommender systems where we observe covariates $x_i\in\R^{d_x}$ for user $i$, $z_j\in\R^{d_z}$ for movie $j$, and dyadic covariates $w_{ij}\in\R^{d_w}$ for the pair $(i,j)$ (for example, how many times the user has watched the movie). Following the approach of~\cite{agarwal2009regression}, we can propose the more flexible generative model
\begin{equation}\label{eq:bigBayes}
  \begin{aligned}
    &\eta_k \simiid N_{d_x}(0,\Sigma), \quad\zeta_k \simiid N_{d_z}(0,\Phi), 
    \quad&k=1,\ldots,r\\
    &U_{ik}\mid \eta_k \simind N(x_i'\eta_k,\sigma^2), \quad V_{jk}\mid \zeta_k \simind N(z_j'\zeta_k,\sigma^2),
    &k=1,\ldots,r\\
    &\Theta_{ij}(\alpha,\beta,\nu,X,Z,W,U,V) = \alpha'x_i + \beta'z_j + \nu'w_{ij} + u_i'v_j &\\
    &Y_{ij} \mid \Theta_{ij} \simind \pi_{\Theta_{ij}}(y),& (i,j)\in\Omega.
  \end{aligned}
\end{equation}
In the above model, $\pi_{\theta}(y)$ represents some model with convex (negative) log-likelihood such as a Gaussian, other exponential family, or log-concave location family. We can also impose a log-concave prior on $(\alpha,\beta,\nu)\in\R^{d_1+d_2+d_3}$ without really increasing the difficulty of the problem, but we treat them as fixed effects for simplicity.

Interestingly, the generalized nuclear norm framework is flexible enough to handle MAP estimation even in this complex model. Again up to a constant shift, the negative log-posterior is
\begin{equation}\label{eq:bigBayesMAP1}
    \cL(\Theta; Y) + \frac{1}{2\sigma^2}\|U-X\eta\|_F^2 
    + \frac{1}{2\sigma^2}\|V - Z \zeta\|_F^2 
    + \frac{1}{2}\|\Sigma^{-1/2}\eta\|_F^2 
    + \frac{1}{2}\|\Phi^{-1/2}\zeta\|_F^2,
\end{equation}
where we have suppressed the dependence of $\Theta$ on the other variables.

The function~\eqref{eq:bigBayesMAP1} may appear daunting at first blush due to the many non-convex bilinear terms. However, by partially minimizing with respect to $\eta$ and $\zeta$ we can massage it into a friendlier form. We first write
\[
\frac{1}{2\sigma^2}\|U-X\eta\|_F^2 + \frac{1}{2}\|\Sigma_\eta^{-1/2}\eta\|_F^2 
  = \sum_k \left(\frac{1}{2\sigma^2}\|U_k-X\eta_k\|_2^2 +
  \frac{1}{2}\|\Sigma_\eta^{-1/2}\eta_k\|_2^2 \right),
\]
which is a separable sum of generalized ridge regression criteria, each regressing $U_k$ against $X$. For any fixed $U_k$, the $k$th term is minimized by setting $\eta_k = (X'X + \sigma^2\Sigma^{-1})^{-1}X'U_k$. Substituting back into the original expression and simplifying, we obtain
\[
\min_{\eta_k} \frac{1}{2\sigma^2}\|U_k-X\eta_k\|_2^2 +
\frac{1}{2}\|\Sigma_\eta^{-1/2}\eta_k\|_2^2 = \frac{1}{2\sigma^2} U_k'(I - H)U_k,
\]
where $H = X(X'X + \sigma^2\Sigma^{-1})^{-1}X'$. After eliminating $\zeta$ the same way, we obtain the criterion:
\begin{equation}
\cL(\Theta; Y) + \frac{1}{2\sigma^2}\|(I-H)U\|_F^2 + \frac{1}{2\sigma^2}\|(I-G)V\|_F^2,
\end{equation}
where $G = Z(Z'Z + \sigma^2\Phi^{-1})^{-1}Z'$. This leads to the equivalent minimization problem
\begin{equation}\label{eq:bigBayesMAP2}
  \cL(\Theta; Y) + \frac{1}{\sigma^2}\|(I-H)\Gamma(I-G)\|_* \sbt \Theta = X\alpha + \beta'Z' + \langle \nu, W\rangle + \Gamma,
\end{equation}
where, we write $\langle \nu, W\rangle = (\nu'w_{ij}: i\leq n, j\leq m)$.

Equation~\eqref{eq:bigBayesMAP2} shows that we are now only penalizing the {\em residuals} of $U_k$ and $V_k$ relative to the (ridge-penalized) linear models in $X$ and $Z$. If desired we can further eliminate the variables $\alpha$ and $\beta$ as discussed in previous sections.

\section{Missing Data and Learning}\label{sec:missing}

In most of the matrix completion literature, the missingness pattern $\Omega$ is implicitly assumed to be uninformative. However, in many of the most salient applications for matrix completion and low-rank modeling, missingness is highly informative. For example, in the case of Netflix data, it is highly implausible that a user chooses movies to watch without any regard to whether they anticipate enjoying those movies. As a result, {\em which movies the users choose to rate} can provide a great deal of insight into their latent types, one of the key insights driving the prize-winning algorithm \citep{bell2007lessons}. We will use the Netflix problem as a running example in this section.

In matrix completion problems, the row and column identities play a very different role than they do in more typical data matrices where each row represents an independent observational unit. Though we abstractly represent these identities by the indices $i$ and $j$, they are not merely anonymous replicates: for example, in the Netflix data they correspond to the identities of the individual users and movies about which we are interested in learning. Thus we consider each entry of the matrix $Y$ to be an observational unit, with possibly unobserved response $Y_{ij}$ and observed predictor variables given by the observed row and column identities $i$ and $j$, as well as any other side information relating to the row, column or entry. Viewed in this way, the parameter matrix $\Theta_{ij}=f(i,j)$ is a regression function mapping the predictors $i$ and $j$ to determine the conditional distribution of the response $Y_{ij}$, and this mapping is parameterized by quantities such as $\alpha_i$, $\beta_j$, $u_i$, and $v_j$, which we view as fixed parameters.

Following convention in the missing-value literature, we can introduce Bernoulli indicator variables for the missingness pattern $M_{ij} = 1\{(i,j)\notin \Omega\}$. For simplicity, we will assume throughout that $(M_{ij},Y_{ij})$ pairs are independent of each other, with
\begin{equation}
  \Xi_{ij} = \log \frac{\P((i,j)\in \Omega)}{\P((i,j)\notin \Omega)},
\end{equation}
or equivalently $\E M_{ij} = (1+\exp\{\Xi_{ij}\})^{-1}$. The data are missing completely at random, then, if $M_{ij}$ is completely independent of $i,j,$ and $Y_{ij}$ --- that is, if every entry is equally likely to be observed. This scenario seems highly unlikely for Netflix data as well as most other matrix completion problems.

\subsection{Data Missing at Random}

The missing data are ignorable in this case if and only if $Y_{ij}$ is independent of $M_{ij}$ given the categorical row and column predictors $i$ and $j$ --- that is, in terms of the framework of~\citet{rubin1976inference}, whether the data are {\em missing at random} (MAR). For example, in the Netflix data, a user may preferentially watch movies that she expects to align with her preferences, but once she decides to watch a movie her decision of whether to rate it is unrelated to her evaluation of the movie.

If the data are MAR in the sense above, then the conditional likelihood of $Y_{ij}$ given $M_i=1$ is the same as the conditional likelihood of $Y_{ij}$ given $M_i=0$. Therefore, we can still learn to predict the missing cases by analyzing the non-missing cases. However, as \citet{bell2007lessons} found, this may be a highly suboptimal, especially if $\Xi_{ij}$ is partly driven by the same parameters $\alpha_i,\beta_j, u_i,$ and $v_j$ that determine $\Theta_{ij}$. 

Viewed this way, missingness at random is a special case of the multi-task learning problem, simply adding more data for estimating the same parameters, possibly in addition to some more parameters. For example, we might add an extra parameter $\rho_i$ for user $i$, parameterizing her propensity to watch more movies, and $\tau_j$ parameterizing movie $j$'s overall prevalence, and model $\Xi_{ij} = \rho_i + \tau_j + u_iv_j'$ or equivalently $\Xi = \rho 1' + 1\tau' + UV'$. We could fit this model with minimal modification to the algorithmic framework described below

Alternatively, we might believe the user/movie interaction should be similar but not exactly the same for $\Xi$ and $\Theta$. Then, we could model
\[
\Xi_{ij} = \rho_i + \tau_j + r_it_j',
\]
and penalize $\lambda_{ru}\|r_i-u_i\|_2^2$ and $\lambda_{tv}\|t_j-v_j\|_2^2$. Mapping this back to a matrix completion problem, we would arrive at partially missing data matrix $\widetilde Y$, and parameter matrix $\widetilde \Theta$, where
\[
\widetilde Y = \begin{bmatrix} Y & \text{---} \\ \text{---} & M \end{bmatrix},
\qquad \widetilde\Theta = \widetilde U \widetilde V' = \begin{bmatrix} U \\ R \end{bmatrix}
\begin{bmatrix} V' & T' \end{bmatrix}.
\]
The character ``---'' denotes completely missing blocks of $\widetilde Y$ that play no role in the likelihood; note this means that the values of $UT'$ and $RV'$ do not matter. Finally, we would be left with the penalized likelihood criterion
\[
\L(\widetilde\Theta; \widetilde Y) + \lambda_{ru}\|U-R\|_F^2 + \lambda_{tv}\|T-V\|_F^2,
\]
where the quadratic penalties can be rewritten as $\|U-R\|_F^2 = \|[1_n 1_n' \;\; -(1_n 1_n')]\widetilde U\|_F^2$. 

More generally, we can imagine an infinitude of possible ways of modeling the missing data pattern, many of which fit quite neatly into the computational framework described here.

\subsection{Data Missing not at Random}

By contrast, we might imagine that another user rates movies strategically, only bothering to rate those movies he especially likes or dislikes, or searching his memory to rate his favorite movies that he watched long ago. In that case, the movies are {\em missing not at random} (MNAR), the most general and least favorable scenario. If the missing mechanism is MNAR then, even if we successfully learn to predict $Y_{ij}$ for observed entries $(i,j)\in \Omega$, we cannot necessarily rely on our predictions to perform well for values of $Y_{ij}$ for the not-yet-observed entries $(i,j)\notin \Omega$. From the perspective of a firm like Netflix, this may pose a major problem, especially if they aim to use the data to recommend movies that a user has not yet rated, but which they believe he would like.

As in most problems, it is impossible to determine from the data alone whether the data are MAR or MNAR (or to correct for MNAR data). To determine whether (or how badly) the data are MNAR, we could however use auxiliary data. For example, we could imagine that Netflix keeps data about (a) which movies a user  watched on Netflix's recommendation, (b) which movies he watched by his own choice (e.g. by searching for them), and (c) which movies he rated without watching on Netflix. Then Netflix could train a model on ratings in groups (b) and (c), and test for group (a); if the predictions are unsuccessful, Netflix could modify its model to take this into account --- for example, by introducing a categorical predictor variable $W_{ij}$ encoding a,b,c, or d if the rating is still missing, and modeling the way that $W_{ij}$ influences $Y_{ij}$.

\section{Computation}\label{sec:computation}

We now review several methods that can be used to solve optimization problems with generalized nuclear norm regularization.

\subsection{Algorithms for solving the nuclear norm regularized problem}\label{sec-nuke-1}
In the discussion below we assume that $\L(\Theta)$ in problem~\eqref{eq:nucLagrange} is convex and differentiable. If $\L(\Theta)$ is nonsmooth (for example, if it corresponds
 to the least absolute deviation loss function) then for a large class of functions, one can use Nesterov's smoothing technique~\citep{nest_05} to smooth the functions and apply the algorithms discussed below.

\subsubsection{Proximal Gradient Algorithms}\label{prox-grad-sec}
 We first discuss obtaining approximate solutions to problem~\eqref{eq:nucLagrange}, minimizing $\cL(\Theta) + \lambda\|\Theta\|_*$. If $\L(\Theta)$ is differentiable with Lipschitz continuous gradient
\[
\| \nabla \L(A) - \L(B)\|_{F} \leq L \| A - B\|_{F}, \quad \forall A, B \in \R^{n\times m},
\]
then problem~\eqref{eq:nucLagrange} is amenable to proximal gradient methods~\cite{fista-09}, which
perform the iterative updates
\begin{equation}\label{prox-update-1}
\begin{aligned}
 \Theta_{k+1} \in& \argmin\limits_{\Theta}~~\frac{L}{2} \| \Theta - ( \Theta_{k} - \frac{1}{L} \nabla \L (\Theta_{k}) ) \|_{F}^2 + \lambda \| \Theta \|_{*} = S_{\lambda/L}  ( \Theta_{k} - \frac{1}{L} \nabla \L (\Theta_{k}) ).
  \end{aligned}
  \end{equation}
In the case where the loss is a least-squares loss over entries in $\Omega$ and $L=1$, this is the {\em Soft-Impute} method of~\citet{mazumder2010spectral}. We note that careful implementations of 
accelerated gradient methods~\cite{fista-09} can also be used for moderate-sized problems -- however, the computations of the subproblems will increase (they will require performing low-rank SVDs of matrices with much larger rank~\citep{mazumder2010spectral}).

Note that if $\lambda$ is large then the number of nonzero singular values in $S_{\lambda}(A)$ is small -- it therefore suffices to compute a low-rank SVD of $A$, which can be significantly cheaper than computing a full SVD of $A$ with cost $O(m^2n)$. If $A$ has special structure for which matrix-vector multiplications of the form $Ab_{1}$ and $A'b_{2}$ are cheap, then a low-rank SVD can be computed with several such matrix-vector multiplications. The popular power method~\citep{GVL83} is often used to compute the largest singular-vector/value of large matrices. The block QR method or alternating least squares~\citep{GVL83,hastie2015matrix} method and algorithms based on Lanczos subspace iterations~\citep{GVL83} as implemented in the PROPACK software~\citep{larsen2004propack} are extremely effective methods for computing the top few singular values and vectors for large matrices for which matrix-vector multiplications are cheap. We discuss some specific cases below.

 \paragraph{Matrix Completion:} For the matrix completion problem with least squares loss, entails computing a low-rank SVD of the matrix 
$P_{\Omega}(Y) + P_{\Omega}^{\perp}(\Theta_{k}),$ (here, $P_{\Omega}(A)$  is a matrix such that its $(i,j)$th entry is $a_{ij}$ for $(i,j)\in \Omega$ and zero otherwise)
which, curiously can be written as the sum of a sparse and low-rank matrix:
\begin{equation}\label{sp-low-rank}
\widetilde{\Theta}_{k} := P_{\Omega}(Y) + P_{\Omega}^{\perp}(\Theta_{k})  =  \underbrace{P_{\Omega}(Y - \Theta_{k})}_{\text{Sparse}}   + \underbrace{\Theta_{k}}_{\text{Low rank}},  
\end{equation}
wherein, $\Theta_{k}$ is anticipated to be of low-rank since a sufficiently large value of $\lambda$ in Problem~\eqref{eq:nucLagrange} encourages a low-rank solution.  In practice, one 
maintains an upper bound $\hat{r}$ on the maximum allowable rank on $\Theta_{k}$ for improved memory and storage usage. In practice, we never need to store 
or form the entire matrix $\Theta_{k}$ -- instead, we store factors $(A_{k}, B_{k})$ where, $\Theta_{k} = A_{k} B_{k}'$. Suppose $r$ is the ``working'' rank of $\Theta_{k}$, i.e., $A_k,B_k$ have $r$ columns each. 
We note that computing $P_{\Omega}(Y - \Theta_{k})=P_{\Omega}(Y) - P_\Omega(\Theta_{k})$ requires 
to evaluate the entries of $\Theta_{k}$  for all $(i,j)\in \Omega$ which can be done with cost $O(|\Omega| r)$ by using the factored representation of $\Theta_{k}$.
Note that multiplying $\widetilde{\Theta}_{k}$ with a vector 
is of cost $O(|\Omega|) +  O((m +n) r)$. Usually in matrix completion problems, we seek $r$ latent factors with  $r \ll m, n$; and 
$|\Omega|$ is comparable to $O((m +n)r)$ -- thus the matrix vector multiplications are of cost $O((m +n) r)$.
When $|\Omega|$ is small, the above techniques also generalize to loss functions corresponding to other members of the generalized linear model family.

\paragraph{Structured Gradients:}

For more general loss functions $\L$, we will need to efficiently compute the gradient of the loss function $\L$ wrt $\Theta$ and also perform fast matrix-vector multiplications 
of the form: $\nabla \L(\cdot) b_{1}$ and $\nabla \L(\cdot)' b_{2}$.
Let us consider a  problem of the form~\eqref{eq:crit2}. 
For methods of Sections~\ref{prox-grad-sec} and~\ref{sec-fw}, one needs to compute the gradient of a smooth function of the form $\nabla_{\Theta} \L(\tilde{P}\Theta \tilde{Q})$ wrt $\Theta$. 
 Writing $\Theta=AB'$ (assuming that $A,B$ have a small working rank) the gradient is given by $\nabla_{\Theta} \L(\tilde{P} \Theta \tilde{Q})=\tilde{P} \nabla_{Z}\L(Z) \tilde{Q}$ with $Z = \tilde{P}\Theta \tilde{Q}$. 
A key factor in computing~\eqref{prox-update-1} requires performing fast matrix-vector multiplications of the form: $\tilde{P} \nabla_{Z}\L(Z) \tilde{Q} b_{1}$ (and also $(\tilde{P} \nabla_{Z}\L(Z) \tilde{Q})' b_{2}$) which is easy to compute as long as:
 Multiplying $\tilde{Q}$  and $\tilde{P}$ with a vector is computationally cheap. This is the case when the matrices $\tilde{P}, \tilde{Q}$ are sparse, low-rank  or the sum of a sparse and low-rank matrix (for example).
All examples discussed in the paper including the ones in Section~\ref{sec:examples} satisfy this property.

\subsubsection{Non-convex Optimization Algorithms}\label{nonconvex-opt-algo}  Another class of algorithms that are commonly used for matrix completion problems directly attempt to optimize the modified problem
\begin{equation}\label{eq:mapRank}
   \minn_{U \in \R^{n \times r},  V \in \R^{m \times r} }  \cL(UV',Y) + \frac{\lambda}{2} \|U\|_F^2 + \frac{\lambda}{2}\|V\|_F^2 
\end{equation}
in terms of the variables $(U,V)$ with $r<\min(n,m)$ using nonlinear optimization methods.

Note the key difference from~Problem~\eqref{eq:map} is that in Problem~\eqref{eq:mapRank} the latent factors $U$ and $V$ are explicitly assumed to have only $r$ columns. As a result, while~Problem~\eqref{eq:map} is always equivalent to~Problem~\eqref{eq:nucLagrange}, Problem~\eqref{eq:mapRank} is not equivalent unless $r$ is larger than the rank of the solution to Problem~\eqref{eq:nucLagrange}. 

\cite{srebro05} propose using gradient decent on formulation~Problem~\eqref{eq:mapRank} for the matrix completion problem. While non-convex problems are prone to local minima and saddle points, it turns out that under certain assumptions~\citep{burer-05,hastie2015matrix,journee2010low}, these non-convex algorithms lead to solutions of the convex optimization problem~\eqref{eq:nucLagrange}.
An intuitive explanation behind this is that~Problems~\eqref{eq:nucLagrange} and \eqref{eq:mapRank} are equivalent for sufficiently large values of $r$. In addition, under some conditions, local minimizers of~Problem~\eqref{eq:mapRank} are global minimizers of Problem~\eqref{eq:nucLagrange}. We note however that certifying whether a pair $(U,V)$ is a local minimizer requires checking the positive semi-definiteness of a Hessian operator, which may be difficult to verify for large scale problems. Furthermore, local minimizers should be distinguished from stationary points and saddle points, which need not correspond to global solutions of the convex problem. We refer the reader to the work of~\cite{hastie2015matrix} for a detailed investigation of these issues  
for the matrix completion problem, wherein the authors also show that non-convex algorithms for the matrix completion problem can be much more efficient than usual proximal gradient type methods for the problem.

\subsubsection{Frank--Wolfe type Algorithms}\label{sec-fw} Recently another class of first order methods known as Frank--Wolfe aka Conditional Gradient  algorithms have gained popularity for nuclear norm regularized problems, especially matrix completion~\citep{frank1956algorithm,jaggi2010simple,freund2015extended}. These methods operate on the constrained version of the nuclear norm regularized problem~\eqref{eq:nucConstrained}. 

The Frank--Wolfe algorithm leads to the update sequence:
\begin{equation}\label{fw-update1}
\Theta_{k+1} = \Theta_{k} + \alpha_{k} ( \tilde{\Theta}_{k+1} - \Theta_{k}) ~~\text{where,}~~ \tilde{\Theta}_{k+1} \in \argmin_{\Theta: \| \Theta \|_{*} \leq \delta } \langle \nabla g(\Theta_{k}), \Theta \rangle~~~~
\end{equation}
for a sequence $\alpha_{k} = 2/(k+2)$. This sequence $g(\Theta_{k})$ converges to the optimum of problem~\eqref{eq:nucConstrained} with a finite time convergence rate of 
$O(1/k)$. An appealing trait of this algorithm is that $\tilde{\Theta}_{k+1}$ (as in~\eqref{fw-update1}) requires computing the largest singular vector/value of the matrix $g(\Theta_{k})$ which can be done via the power-method, for example. For matrix completion, $\nabla g(\Theta_k) = P_{\Omega}( Y - \Theta_{k})$ which is a sparse matrix with $O(|\Omega|)$ nonzero entries and hence a power method
will entail a per-iteration cost of $O(|\Omega|)$. In case $g(\Theta_k)$ has no specialized structure, computing $\tilde{\Theta}_{k+1}$ can be achieved via the power method with 
cost $O(mn)$  -- this is usually much cheaper than computing a thresholded SVD as in~\eqref{prox-update-1}. The caveat of the Frank--Wolfe method is that 
it can take several iterations to reach an approximate solution to problem~\eqref{eq:nucConstrained} with a small rank (even if we assume that the problem admits a low-rank solution).
For example, if we adjust $\delta$ such that the rank of $\hat{\Theta}$ (an optimal solution to problem~\eqref{eq:nucConstrained}) is 20, say, $\rnk(\Theta_k)$ can quite easily become of the order of a thousand with as many iterations -- as the number of iterations increase, the rank gradually decreases. This is in contrast to proximal gradient methods~\citep{mazumder2010spectral,hastie2015matrix} where, the nuclear norm thresholding operator induces a low-rank solution.
There are sophisticated variants of the Frank--Wolfe
method that can address these shortcomings, with additional computational cost --- we refer the reader to~\cite{freund2015extended} and references therein for an in-depth investigation.

\subsection{Solving the Generalized Nuclear Norm Problem}\label{sec:gen-nuke}
Motivated by problem~\eqref{eq:genNuc}, let us now discuss how to minimize
$\L(\Theta; Y) + \lambda \| P \Theta Q\|_{*}$,
depending upon choices of $P,Q$. 

If $P,Q$ are invertible, then problem~\eqref{eq:genNuc} with a suitable change of variables can be reformulated as an instance of problem~\eqref{eq:nucLagrange} -- and the methods described above apply. 
This approach is suitable as long as performing  the matrix inversions for $P,Q$ are computationally feasible (for instance $m, n$ of the order of a few thousands each).
Even if $m,n$ are large (in the order of tens of thousands), it may be easy to  invert $P$ especially if it has some special structure. For example, if 
$P = I-H$ for $H$ low-rank, then $P^{-1}$ (assumed to exist) can be obtained by using the Sherman Woodbury formula. A similar story applies to $Q$ as well. 
In addition to be able to compute the inverses of $P,Q$; one also needs to compute the gradient quite efficiently -- this is possible as long as it is fast to multiply $P^{-1},Q^{-1}$  with vectors.

If at least one of $P$ or $Q$ is low rank, then problem~\eqref{eq:genNuc} can be solved quite easily, using the Alternating Direction Method of multipliers method~\citep{boyd-admm1} as we discuss in Appendix~\ref{sec-solve-ref11}. 

Finally, another approach to solve problem~\eqref{eq:genNuc} is to express it in the form of problem~\eqref{eq:crit1} in terms of the latent factors $U,V$ (where, we assume 
$\Theta = UV'$) and directly apply 
nonlinear optimization methods on the problem -- for example, one can apply gradient descent on the function with respect to latent variables $U_{n \times r},V_{m \times r}$. A stationary point of such an 
algorithm will correspond to the minimum of the convex problem~\eqref{eq:genNuc} if: $r$ is chosen sufficiently large, and the point $\hat{\Theta}  = \hat{U} \hat{V}'$ corresponds to 
a local minimum of the objective function -- this usually entails checking if $\hat{U} \hat{V}'$ satisfies the optimality condition of the corresponding convex problem. 

\paragraph{Multiple Blocks:} Let us consider the general convex problem~\eqref{eq:bigBayesMAP2}, where, 
$\Theta_{ij} =  X\alpha + \beta'Z' + \langle \nu, W\rangle + \Gamma$. 
 We can apply a block coordinate descent algorithm~\citep{bertsekas-99} across the blocks 
 $(\alpha, \beta, \nu)$ and $\Gamma$. The optimization problem wrt the block $\Theta^{(1)}$ is a simple convex problem and 
 can be solved quite easily using standard methods. The optimization problem wrt $\Gamma$ is a (generalized) nuclear norm regularized problem, and has been discussed above.

We note that the candidates described above are all competitive methods and are likely to yield comparable performances on several problem instances. For specialized implementations and improved performance, 
the above possibilities need to be examined carefully based on problem structure, size of the problem, and the particular datasets under study. For example, for the matrix completion problem with nuclear norm regularization and squared error loss, a vanilla implementation of
the proximal gradient method (Section~\ref{prox-grad-sec}), the Frank-Wolfe method (Section~\ref{sec-fw}) or the non-convex optimization based algorithm (Section~\ref{nonconvex-opt-algo}) will have comparable performance -- for more efficient computational performance, specialized implementations building and extending these basic algorithmic principles 
are needed --- see for example~\cite{hastie2015matrix,freund2015extended}.

\section{Data Examples}\label{sec:examples}

We now describe the results of our method on two applications, meant
both to illustrate the scalability of our algorithm, and to suggest
the level of generality of possible models by way of example.

\subsection{Functional Data Reconstruction}\label{sec:functional}

Our second example is of a more nonparametric flavor and involves real data.  We begin with
200 noisy functions measured at 256 equally spaced frequency points, representing measured log-periodograms of several phonemes spoken by various subjects.  This data set was analyzed by \citet{hastie1995penalized} to demonstrate a variant of discriminant analysis with smoothness penalty applied.

We artificially construct a sparsely-observed data set by sampling each curve at 26 random points, and set ourselves the objective of reconstructing the functions based on these relatively few samples, exploiting smoothness and a low-rank assumption about the curves.

We impose a simple model on the column variables $v_j$; they are constrained to lie in the natural spline basis with 12 degrees of freedom, and we shrink them toward the natural spline basis with 4 degrees of freedom.  We estimate the principal components analysis model with unpenalized marginal column (time) effects.
\begin{align}
  \text{min. } &\cL(\Theta;Y) + \lambda_\Gamma\|\Gamma_{ij}^{(2)}\|_*\\
  \text{s.t. } &\Theta_{ij} = \mu + \beta_j + \Gamma_{ij}^{(1)}H_4 + \Gamma_{ij}^{(2)}H_{12},
\end{align}
where $H_d$ represents a $d$-dimensional natural spline basis (with intercept). This can be framed (somewhat artificially) as a Bayesian prior on $V$ with flat variance in the directions of $H_4$, finite positive variance in the directions of $H_{12}$, and zero variance on other directions.  Although this data set is relatively small, it can be computationally advantageous to constrain $\Gamma$ as we have done here, since it reduces the size of our optimization variables (e.g. $\Gamma^{(2)}$ is $n \times 12$ instead of $n \times
256$). We compare our proposed method to matrix completion using the standard nuclear norm, which does not exploit smoothness.  The right panel of Figure~\ref{fig:phoneme} shows that side information cuts MSE by a sizeable fraction.

\captionsetup[subfigure]{width=1.2\textwidth}
\begin{figure}
  \centering
  \begin{subfigure}[t]{0.5\textwidth}
    \includegraphics[width=\textwidth]{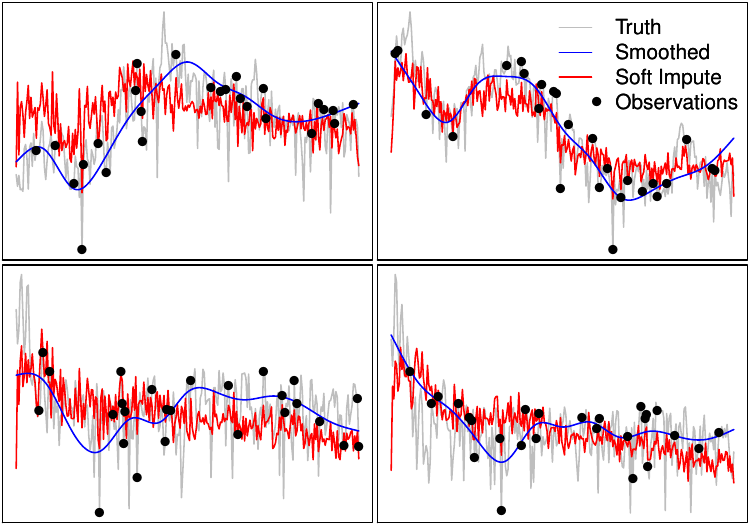}
    \caption{Reconstruction for the first four phoneme curves, with regularization parameter $\lambda$ chosen by cross-validation.}
    \label{fig:phonemeRecon}
  \end{subfigure}
  \hspace{.1\textwidth}
  \begin{subfigure}[t]{0.34\textwidth}
    \includegraphics[width=\textwidth]{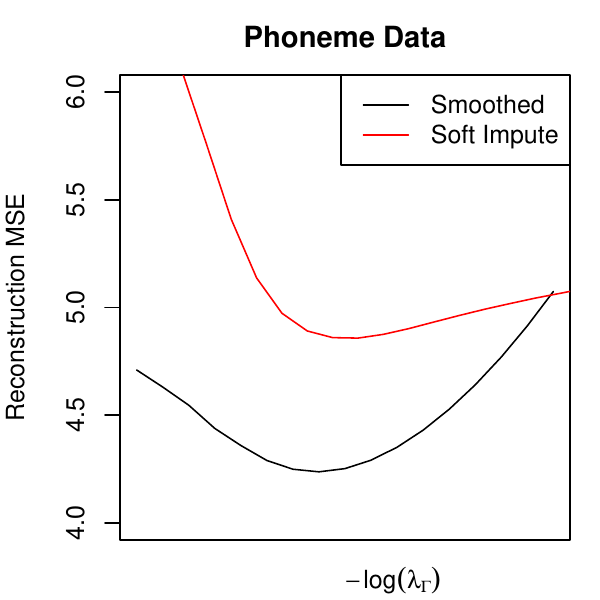}
    \caption{Mean squared error for held-out data as a function of $\lambda$.}
    \label{fig:phonemeMSE}
  \end{subfigure}
  \caption{Phoneme data, comparison of Soft-Impute (matrix completion with the standard nuclear norm penalty) against a generalized nuclear norm penalty with smoothness enforced.}
\label{fig:phoneme}
\end{figure}

\subsection{Reduced-Rank Poisson Regression for Ecological Modeling}

As a third example to illustrate the richness of our modeling framework, we consider an ecological application, species distribution modeling using presence-only data. On a geographic domain $\mathcal{D}$, we observe a process of point observations for each of $m$ species, with the goal of determining the abundance of each species as a function of geographic location, or understanding the determinants of habitat suitability. The observations typically arise from opportunistic sources such as museum collections or citizen science data, leading to a strong sampling bias toward population centers. Our model extends a model proposed in \citet{fithian2014bias} by introducing low-rank regularization to borrow strength across species.

For $s\in\mathcal{D}$, let $x(s)$ denote a $d$-variate vector of habitat covariates that drive species abundance, and let $z(s)$ denote other covariates driving the sampling bias. We assume that species $j$ has a latent {\em species process} $\mathcal{S}_j$ representing all locations where species $j$ occurs, modeled as an inhomogeneous Poisson point process with {\em species intensity} $\nu_j(s) = \exp\{\alpha_j + x(s)'\beta_j\}$. The species process is then filtered through a biased observation model wherein each occurrence is observed with probability $b_j(s)$, so that we only observe the thinned Poisson process $\mathcal{T}_j$ with intensity $\nu_j(s)b_j(s)$. We model $b_j(s)= \exp\{\epsilon_j + z(s)'\delta\}$ (with $\delta$ {\em not} varying by species), reflecting an opinion that the spatial bias is a function of observer behavior only. 

Typically the geographic domain is discretized into $n$ pixels reflecting the resolution of covariate measurements; if $s$ represents a pixel with unit area, let $Y_{sj}$ denote the number of $s\in\mathcal{T_j}$ falling into pixel $s$. Then,
\[
Y_{sj} \sim \text{Pois}\left(\exp\{\alpha_j + x(s)'\beta_j + \epsilon_j + z(s)'\delta\}\right).
\]
Since $\alpha_j$ and $\epsilon_j$ are unidentifiable in this model, the species intensity $\nu_j$ is also unidentifiable. However, we can estimate the normalized {\em species distribution} $p_j(s) = \nu_j(s) / \int_{\mathcal{D}} \nu_j(s)$, which depends only on $\beta_j$ and is therefore identifiable.

To borrow strength across species, we can model $B=UV'$, where $B = [\beta_1 \cdots \beta_m]$, and $U \in \R^{d \times r}, V\in\R^{m \times r}$. In effect we are positing that $r$ latent habitat covariates $w(s) = x(s)'U$ can capture all of the important signal, with $V' = [v_1 \cdots v_m]$ representing the effect of the latent covariates on each species. Replacing the non-convex rank constraint with a nuclear norm penalty, we arrive at a {\em reduced-rank Poisson regression} objective:
\begin{equation}\label{eqnEcol}
\minn_{\alpha,\Theta,\delta} \L(\Pi_X\Theta + 1\zeta' + \delta'Z1', Y) + 
\lambda\|\Theta\|_*,
\end{equation}
where $\Pi_X$ denotes projection onto the column space of $X$ and $\zeta_j = \alpha_j + \epsilon_j$). Note that while this application is not posed directly as matrix completion as such, estimating $p_j(s)$ essentially amounts to predicting the locations of the missing observations.

\begin{figure}
  \centering
  \begin{subfigure}[t]{0.45\textwidth}
    \includegraphics[width=\textwidth]{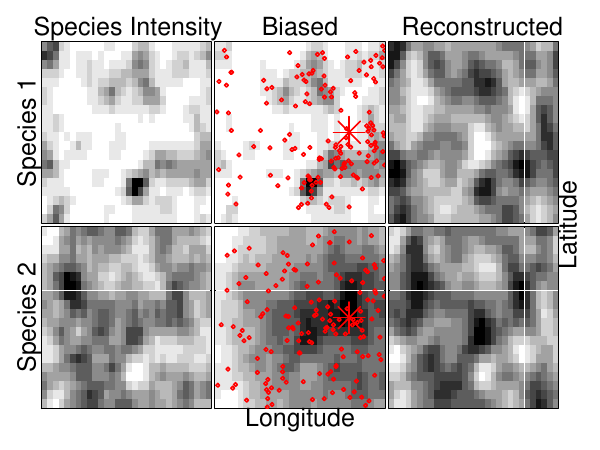}
    \caption{Top row: The left panel shows $\nu_1$, the species intensity. The middle panel shows $\nu_1b_1$, the biased intensity, with the observed process $\mathcal{T}_1$ plotted as red circles. The red star denotes the location of the ``town'' which drives the bias. The right panel shows the reconstructed species distribution $\hat p_1$, for the value of $\lambda$ selected by cross-validation. Bottom row: Same quantities for species 2.}
    \label{fig:ecolIntensity}
  \end{subfigure}
  \hspace{.1\textwidth}
  \begin{subfigure}[t]{0.39\textwidth}
    \includegraphics[width=\textwidth]{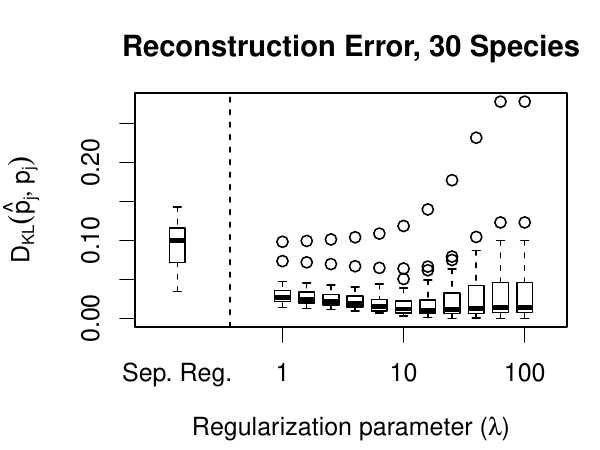}
    \caption{Comparison of the reconstruction error for the separate-regressions method versus the regularized method for a range of $\lambda$ values. For each fit, the box summarizes $D_{KL}(\hat p_j, p_j)$ for $j=1,\ldots,30$. The separate-regressions method performs quite poorly compared to the regularized method.}
    \label{fig:ecolResults}
  \end{subfigure}
  \caption{Results for ecological simulation.}
\label{fig:phoneme}
\end{figure}

We illustrate this method on simulation data, where the ground truth is known and estimation accuracy can be directly measured. On a $20\times 30$ grid of pixels in the unit square $\mathcal{D}=[0,1]\times[0,1]$, we generate $d=30$ covariates $x(s)$ as moving-average Gaussian processes. We then randomly generate latent factors $U$ and species loadings $V$ for each of $m=30$ species, populating both matrices with i.i.d. Gaussian random variables. Next, we plant a ``town'' at location $s^*=(0.8,0.5)$ and let $z(s) = -\|s - s^*\|_2^2$. Finally, we set $\alpha_j$ to normalize the intensities so that $\int_{\mathcal{D}} \nu_j(s)b_j(s) = 150$ for each species. Figure~\ref{fig:ecolIntensity} shows the species intensity, biased intensity, and reconstructed species distribution for the first two species, for the best-performing value of $\lambda$ on a validation set.

\captionsetup[subfigure]{width=1.2\textwidth}

We compare our regularized estimator to a simpler method wherein we estimate $\beta_j$ for $j=1,\ldots,m$ via a separate log-linear Poisson regression for each species. To simplify estimation, the separate-regressions method is given perfect {\em a priori} knowledge of $\delta$, so $\delta$ need not be estimated. Even with this advantage, the separate-regressions method overfits badly; 150 observations are not enough to accurately estimate a density with 30 parameters. Figure~\ref{fig:ecolResults} shows a boxplot of the Kullback-Leibler distance $D_{KL}(\hat p_j \| p_j)$ for each value of $\lambda$, and for the separate regressions method. The full simulation code is given in the supplementary materials.

\section{Discussion}

We presented a framework for scalable convex optimization on matrix
completion problems incorporating side information.  The information
can be diverse in its source, as long as it can be represented
ultimately as some quadratic penalty which can be applied or inverted
with ease.  We have seen two examples where side information of
different kinds is advantageous for predictive performance.

Although the bottleneck in our algorithm is an SVD of a large
matrix, we can attain rapid convergence by exploiting the structure of
the SVD target, which is often easy to apply.

\section*{Acknowledgments}
A previous unpublished version of this manuscript is available on arxiv~\citep{fithian2013scalable}. William Fithian was supported in part by National Science Foundation VIGRE grant DMS-0502385 and the Gerald J. Lieberman Fellowship. 
R. Mazumder is partially supported by ONR Grant N000141512342.
We are grateful to Trevor Hastie and Julie Josse for helpful discussions.

\clearpage

\bibliographystyle{plainnat}

\bibliography{biblio,rahul_dbm4.bib,rahul_dbm3}

\begin{appendix}

\section{Proof of Theorem~\ref{propProj}}\label{sec:proof-propProj}
\begin{proof}
  Let $\Pi_P$ and $\Pi_Q$ denote projections onto the images of
  $P$ and $Q$.  Then
  \begin{align}\label{eqnProof1}
    & \inf_\Theta ~~  \cL(\Theta) + \|P\Theta Q\|_*\\
    &= \lim_{\ep\downto 0}~ \inf_\Theta~~ \cL(\Theta) + \|(P+\ep \Pi_P^\perp)\Theta
    (Q+\ep \Pi_Q^\perp)\|_*\\\label{eqnProof2}
    &= \lim_{\ep\downto 0}~ \inf_{U,V}~~ \cL(UV^T) + \frac{1}{2}\|(P+\ep
    \Pi_P^\perp)U\|_F^2 + \frac{1}{2} \|(Q+\ep \Pi_Q^\perp)V\|_F^2\\\label{eqnProof3}
    &= \inf_{U,V}~~ \cL(UV^T) + \frac{1}{2}\|PU\|_F^2 + \frac{1}{2} \|QV\|_F^2
  \end{align}
  We can see \eqref{eqnProof2} by changing variables to
  $\widetilde\Theta = (P+\ep \Pi_P^\perp)^{-1}\Theta(Q+\ep
  \Pi_Q^\perp)^{-1}$, $\widetilde
  U = (P+\ep \Pi_P^\perp)^{-1}U$, and $\widetilde V = (Q+\ep
  \Pi_Q^\perp)^{-1}V$.
  \eqref{eqnProof1} and \eqref{eqnProof3} follow from the fact that
  $\|\Theta\|_* \geq \|\Pi \Theta\|_*$ for any
  projection $\Pi$ (in this case $\Pi_P$),
  and similarly $\|U\|_F \geq \|\Pi U\|_F$.

  For any $\Theta$ we can find
  $\Theta_1,\Theta_2,\Theta_3$ for which  ${\Theta=
    P^+\Theta_1 Q^+ + \Pi_P^\perp\Theta_2 +
    \Theta_3\Pi_Q^\perp}$.  Then
  \begin{align}\label{eqnThetaDecomp1}
    & \inf_{\Theta} \cL(\Theta) + \|P\Theta Q\|_* \\
    &= \inf_{\Theta_1,\Theta_2,\Theta_3} \cL(P^+\Theta_1 Q^+ + \Pi_P^\perp\Theta_2 +
    \Theta_3\Pi_Q^\perp) + \|\Pi_P \Theta_1 \Pi_Q\|_* \\ \label{eqnThetaDecomp}
    &= \inf_{\Theta_1,\Theta_2,\Theta_3} \cL(P^+\Theta_1 Q^+ + \Pi_P^\perp\Theta_2 +
    \Theta_3\Pi_Q^\perp) + \|\Theta_1\|_*
  \end{align}
  \eqref{eqnThetaDecomp1} holds because ${P(P^+\Theta_1 Q^+ + \Pi_P^\perp\Theta_2 +
    \Theta_3\Pi_Q^\perp)Q = \Pi_P\Theta_1\Pi_Q}$. \eqref{eqnThetaDecomp}
  holds because $\|\Pi_P \Theta_1 \Pi_Q\|_*\leq \|\Theta_1\|_*$ and we
  can attain the minimum by replacing $\Theta_1$ with
  $\Pi_P\Theta_1\Pi_Q$, which does not change the $\cL(\cdot)$ term.
\end{proof}

\section{Solving Problem~\eqref{eq:genNuc} when $P$ is low rank}\label{sec-solve-ref11}
Assuming wlog that $P$ is low-rank with SVD $P = UDU'$ (with $D$ a $J\times J$ diagonal matrix) the problem can be reformulated as:
\begin{equation}\label{ref-11}
\min_{\Gamma, \Theta} ~~ \cL(\Theta; Y) + \lambda \| \Gamma \|_{*}  ~~ \sbt ~~ DU'\Theta Q = \Gamma.
\end{equation}
The above problem where $\Gamma$ is a wide matrix with low rank (at most the rank of $P$) can be solved quite easily with a splitting method with the ADMM method~\cite{boyd-admm1}.
A simple application of the ADMM procedure leads to problem~\eqref{ref-11}:
$$  H(\Gamma, Z) =  \cL(\Theta; Y) + \lambda \| \Gamma \|_{*} + \langle Z,  DU'\Theta Q -  \Gamma\rangle   + \frac{\rho}{2} \| DU'\Theta Q -  \Gamma\|_{F}^2. $$
Note that $\Gamma$ is a low rank rectangular matrix $\Gamma \in \R^{ J \times m}$ with $J$ small. 
The ADMM procedure requires optimizing 
$ H(\Gamma, Z) $ wrt $\Gamma$ for $Z$ fixed --- this can be achieved easily using a proximal gradient method --- the nuclear norm thresholding operation can be 
performed quite easily since it requires the SVD of a low-rank rectangular matrix (same dimension as $\Gamma$). The update step 
$\min_{Z} H(\Gamma, Z)$ with $\Gamma$ held fixed can be achieved by solving a simple (unconstrained) convex function by performing gradient descent, for example.

We note that this problem can also be solved by using Proposition~\ref{propProj}; and using proximal gradient descent methods on the reformulated problem of the form~\eqref{eq:crit2}.
\end{appendix}

\end{document}